\documentclass[sigconf]{acmart}

\AtBeginDocument{%
  }

\setcopyright{acmlicensed}
\copyrightyear{2025}
\acmYear{2025}
\setcopyright{cc}
\setcctype{by}
\acmConference[IVA Adjunct '25]{ACM International Conference on Intelligent Virtual Agents}{September 16--19, 2025}{Berlin, Germany}
\acmBooktitle{ACM International Conference on Intelligent Virtual Agents (IVA Adjunct '25), September 16--19, 2025, Berlin, Germany}\acmDOI{10.1145/3742886.3756719}
\acmISBN{979-8-4007-1996-7/2025/09}

\acmSubmissionID{35}

\begin{document}

\title{Evaluation of a Sign Language Avatar on Comprehensibility, User Experience \& Acceptability}

\author{Fenya Wasserroth}
\orcid{0009-0001-8292-6522}
\affiliation{
  \institution{Technische Universität Berlin}
  \city{Berlin}
  \country{Germany}
}
\email{wasserroth@campus.tu-berlin.de}

\author{Eleftherios Avramidis}
\orcid{0000-0002-5671-573X}
\affiliation{
  \institution{German Research Center for AI}
  \city{Berlin}
  \country{Germany}
}
\email{eleftherios.avramidis@dfki.de}

\author{Vera Czehmann}
\orcid{0009-0004-8007-6503}
\affiliation{
  \institution{German Research Center for AI}
  \city{Berlin}
  \country{Germany}
}
\affiliation{
  \institution{Technische Universität Berlin}
  \city{Berlin}
  \country{Germany}
}
\email{vera.czehmann@dfki.de}

\author{Tanja Kojic}
\orcid{0000-0002-8603-8979}
\affiliation{%
  \institution{Technische Universität Berlin}
  \city{Berlin}
  \country{Germany}
}
\email{tanja.kojic@tu-berlin.de}

\author{Fabrizio Nunnari}
\orcid{0000-0002-1596-4043}
\affiliation{
  \institution{German Research Center for AI}
  \city{Saarbrücken}
  \country{Germany}
}
\email{fabrizio.nunnari@dfki.de}

\author{Sebastian Möller}
\orcid{0000-0003-3057-0760}
\affiliation{
  \institution{Technische Universität Berlin}
  \city{Berlin}
  \country{Germany}
}
\affiliation{
  \institution{German Research Center for AI}
  \city{Berlin}
  \country{Germany}
}

\email{sebastian.moeller@tu-berlin.de}

% \renewcommand{\shortauthors}{Wasserroth, Avramidis et al.}

\begin{abstract}
% Advancements in augmented reality (AR) and assistive technologies offer unprecedented opportunities to enhance communication for the deaf and hard of hearing. 
This paper presents an investigation into the impact of adding adjustment features to an existing sign language (SL) avatar on a Microsoft Hololens 2 device. Through a detailed analysis of interactions of expert German Sign Language (DGS) users with both adjustable and non-adjustable avatars in a specific use case, this study identifies the key factors influencing the comprehensibility, the user experience (UX), and the acceptability of such a system. 
Despite user preference for adjustable settings, no significant improvements in UX or comprehensibility were observed, which remained at low levels, amid missing SL elements (mouthings and facial expressions) and implementation issues (indistinct hand shapes, lack of feedback and menu positioning). 
Hedonic quality was rated higher than pragmatic quality, indicating that users found the system more emotionally or aesthetically pleasing than functionally useful. 
Stress levels were higher for the adjustable avatar, reflecting lower performance, greater effort and more frustration. Additionally, concerns were raised about whether the Hololens adjustment gestures are intuitive and easy to familiarise oneself with.
While acceptability of the concept of adjustability was generally positive, it was strongly dependent on usability and animation quality. 
This study highlights that personalisation alone is insufficient, and that SL avatars must be comprehensible by default. Key recommendations include enhancing mouthing and facial animation, improving interaction interfaces, and applying participatory design.
% Findings indicate that the ability to personalize avatars significantly enhances comprehensibility and enriches user experience, highlighting the critical need for adaptive user interfaces in assistive technologies. 

% The research provides valuable insights into designing digital communication tools that are sensitive to diverse user needs, advocating for the integration of continuous user feedback in the development process to ensure functionality and inclusivity.
\end{abstract}


\begin{teaserfigure}
\centering
  \includegraphics[width=.5\textwidth]{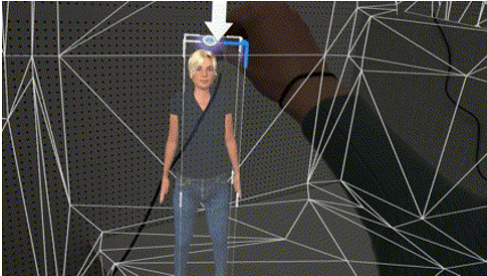}
  \includegraphics[width=0.43\textwidth]{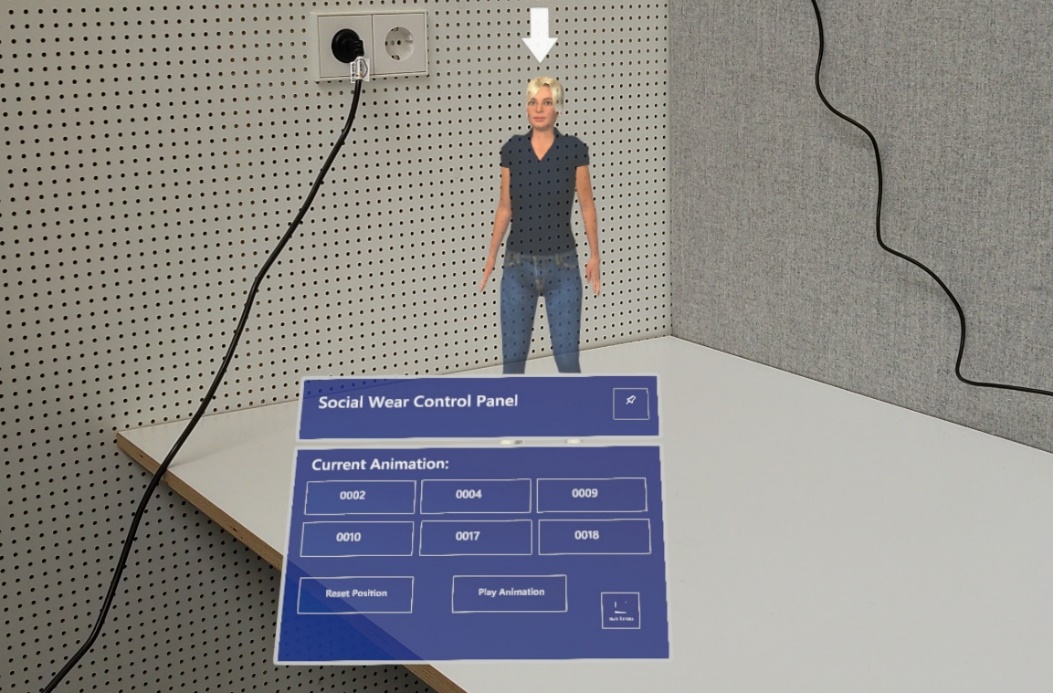}
  \caption{Avatar modification example}
  \Description{Avatar modification example.}
  \label{fig:teaser}
\end{teaserfigure}


\maketitle
\section{Introduction}\label{introduction}

Over 70 million people worldwide are deaf and use more than 200 national sign languages (SLs) \cite{WFD2021}. For many, SL is their first and preferred language, and written text remains a less accessible second language. SL interpreters thus support communication between hearing individuals and deaf people. According to the Federal Association of Sign Language Interpreters in Germany, around 850 interpreters serve approximately 80,000 hearing-impaired individuals in Germany \cite{bgsd2020}. As a result, there are everyday situations where no interpreter is available, complicating communication and limiting access—for example, to train announcements.

SL avatars could help improve accessibility, making research in this area crucial. While one might argue displaying text subtitles on AR glasses could be sufficient, this overlooks the cultural and linguistic significance of SLs. Prioritising text over SL in assistive technologies risks further marginalising SL users and perpetuating a hearing-centric view of communication. If successful, communication barriers between deaf and hearing communities could be reduced long-term. These avatars are not intended to replace human interpreters, but to support the autonomy of deaf individuals in situations where interpreters are unavailable \cite{Krausneker2022}.

At the same time, \emph{augmented reality} (AR) technology is evolving, finding applications in fields like surgery \cite{Kim2017} and automotive displays. AR can embed artificial information into the real environment—an advantage that can benefit SL avatars \cite{Yang2022}. %This allows for avatars to appear next to a speaker, for example \cite{Nguyen2021}.
%Designing supportive, unobtrusive systems is essential \cite{DIN9241_210}. \emph{Human-centred design} (HCD) aligns systems with user needs \cite{DIN9241_210}, making the involvement of the target group in development vital \cite{DIN9241_210}. 
In research led by hearing individuals, expertise from the deaf and hard-of-hearing community is particularly important \cite{Fox2023}.

This study examines how avatar adjustments—such as scalable size and adjustable spatial positioning—affect comprehensibility, user experience, and acceptability in an AR SL assistant. A user study with a mixed-method approach was conducted that combined quantitative data from standardised questionnaires with qualitative responses from an interview with the participant. The following research questions guide the investigation:

\begin{itemize}
\item \textbf{Comprehensibility}: Can adjustment options of a SL avatar improve the comprehensibility of the signed sentences?

\item  \textbf{User experience}: Do adjustment options contribute to a
better user experience of a SL avatar?

\item  \textbf{Acceptability}: To what extent do the adjustable setting options
contribute to the acceptability of a SL avatar? Are there
differences in acceptability and, if so, can factors be identified that
explain these differences?
\end{itemize}

\section{Related work}
\subsection{Sign language avatars}\label{sign-language-avatars}

Research in this area focuses on three core themes: acceptance, qualitative user requirements, and technical development. 

Qualitative requirements include both the technical necessities of transmitting SL and the needs of the deaf and hard-of-hearing community. SL involves hand shapes, movement, location, facial expressions, and lip movements. These components must be animated and synchronised, and signs combined into fluid sentences \cite{Bernhard2022,Krausneker2022}. Facial expressions and lip movements are crucial for understanding and emotional expression \cite{Kipp2011,Lacerda2023}.
Presentation and control aspects are also being studied. Avatars may have comic-like or human-like appearances \cite{Aziz2023}, though comic styles can be situationally appropriate \cite{Kipp2011}. More human-like avatars risk entering the "uncanny valley", reducing acceptance \cite{Mori2019,Bragg2019}. Personalisation is valued, but customisations that improve comprehension—like size or speed—are considered more important than appearance \cite{Nolte2023,Kipp2011}.
Technically, avatars must convert spoken language into signed output. This involves automated translation and/or avatar synthesis—challenging, because SL is neither linear nor written. Methods involving motion capture (MoCap) or glosses are common but limited \cite{Wolfe2023}. Precise synchronisation of visual elements is required \cite{Wolfe2023,Aziz2023}. %However, storing all nuances of sign language in MoCap databases is unrealistic \cite{Bragg2019}.

The use of Augmented Reality for SL avatars has been suggested in order to allow deaf or hard of hearing people to see an interpreter next to the speaking person or other visual information simultaneously, without having to switch their gaze \cite{nguyen-etal-2021-evaluating,nguyen-etal-2021-automatic,luo-etal-2022-avatar}. Many efforts have focused on educational use cases \cite{kozuh-etal-2015-assessing, soogundjoseph-2019-signar, quandt-2020-teaching, aditama-etal-2021-designing, cabanillas-carbonell-etal-2022-mobile, yang-etal-2022-holographic, liang-etal-2024-iknowisee}, most of them in prototypical stage with either no or only limited user evaluation. 
Among these, \cite{yang-etal-2022-holographic} contains a user study similar to ours, showing that users prefer a full-body vs. a half-body signing AR avatar, with the possibility to adjust its position. 

%double blind?
%\subsubsection{The DFKI sign language avatar}\label{the-dfki-sign-language-avatar}

The DFKI avatar for DGS~\cite{nunnari-etal-2023-incorporating}, on which this work is based (Figure~\ref{fig:teaser}), is an animated AR SL interpreter currently running on Microsoft HoloLens 2. The version used for this experiment can sign six pre-recorded sentences related to public transport, developed within the AVASAG research project~\cite{Nunnari2021}. Users can start animations via a menu after a brief loading time.
The avatar’s size, 3D position, and orientation are adjustable. A visible frame appears when the user’s hand is nearby. Users can scale the avatar by dragging a corner, and rotate it using side handles. %, or move it by gripping and placing it elsewhere in the room.

\subsection{Comprehensibility}\label{comprehensibility}

To investigate the understanding of natural language, the different concepts must first be defined and differentiated from one another. The following levels are distinguished: Understandability, comprehensibility, communicability and comprehension \cite{Moeller2017}. Understandability describes the lowest level, as just simply the ability to transmit the signal \cite{Moeller2017}, and can apply to units of various sizes. Comprehensibility describes the identification of the signal based on its form \cite{Moeller2017}. It is influenced by the comprehensibility of individual units and by lexical, syntactic and semantic context \cite{Moeller2017}. When applied to SL, comprehensibility describes how well the content of a signed statement can be identified \cite{Crowe2019}. Communicability describes whether an utterance is understood in the way it was intended \cite{Moeller2017}. In addition to comprehensibility, this is influenced by meta-factors such as delay times of the signal. A successful communication process ends in understanding \cite{Moeller2017}.

Various metrics are used to evaluate the quality of the animated SL avatars. Metrics from machine translation (MT) such as the word error rate (WER), the BiLingual Evaluation Understudy (BLEU) or NIST, a further development of BLEU \cite{Kahlon2023}, are frequently used. These metrics, however, were designed to assess precision by comparing output to a human translation \cite{Kahlon2023, Papineni2002}. While well established for machine text-to-text translations, there is no standardised method of evaluating text-to-SL translations \cite{Mueller2023}. Various forms of sign error rate (SER), such as the multiple reference sign error rate (mSER), are also used in the evaluation of SL \cite{LopezLudena2012}. All of these metrics measure comprehensibility, but do not yet provide any information about how comprehensible it is for users.

Martino et al.\ \cite{Martino2017} assessed the comprehensibility of individual signs using an "isolated sign comprehensibility test" (ISIT), in which the animation is evaluated against the video of a SL interpreter. Nevertheless, Crowe et al.\ \cite{Crowe2019} report a lack of studies on the comprehensibility of SL from the recipient’s perspective, which is an indicator that there is no standardised method. However, the importance of comprehensibility for SL avatars has been researched: especially in important areas of application, where errors in comprehensibility have critical consequences, full comprehensibility must be guaranteed regardless of the SL level \cite{Krausneker2022}. Crowe et al.\ \cite{Crowe2019} investigated which factors influence the comprehensibility of SL. The use of grammatical features and the clarity of signs are considered to promote comprehensibility, whereas the exclusive use of the finger alphabet leads to lower comprehensibility.

On the question of what influence comprehensibility has on the acceptance and evaluation of SL avatars, Wolfe et al.\ \cite{Wolfe2022} predict a very large influence, suggesting that judgement is dependent on the translation quality and the associated effort of comprehension. In a study with focus groups, however, it was observed that the participants did not differentiate between the pure translation quality and the animation \cite{Krausneker2022}. Nevertheless, these results show that the comprehensibility of animated signs plays a key role in the research and development of SL avatars. In addition, the study by Smith and Nolan \cite{Smith2016} suggests that the appearance of the avatar affects comprehensibility too. For example, a human-like avatar achieved a better comprehensibility rate than a comic-like one.

\subsection{Acceptability}\label{acceptability}

Acceptability describes the proportion of the target group that would actually use a SL avatar \cite{Moeller2017}. Therefore, only the preliminary stage of acceptability can be observed by the specified use in the context of the evaluation \cite{Alexandre2018}.
There is a wide range of models and approaches in the literature to describe factors of acceptability. A well-known model is the Technology Acceptance Model (TAM), which is based on perceived \emph{usefulness} and \emph{ease of use} \cite{Davis1989}. \cite{Othman2024} extended this to TAMSA, adding the factor \emph{trust}, due to lower trust in avatars compared to human interpreters. TAMSA, however, was developed with Deaf individuals in Qatar exclusively, limiting generalisability \cite{Othman2024}.
Acceptability often depends on use context~\cite{Kipp2011,Krausneker2022}. Nielsen’s model distinguishes social from practical acceptability \cite{Nielsen1993}. Social acceptability is influenced by context, norms, and experience, while practical acceptability covers aspects like usefulness and reliability.
Social acceptability improves when avatars are intended to complement interpreters in otherwise inaccessible contexts \cite{Krausneker2022}. Independence and flexibility enhance perceived usefulness. Voluntary use is also key \cite{Kipp2011}. Individual background influences acceptance -- people who learned SL early tend to show less interest in avatars~\cite{Quandt2022}.

Technical challenges must be addressed to meet user expectations. Including deaf individuals in the development process provides essential expertise and fosters trust \cite{Krausneker2022,Picron2024}. Engaging users in development-related focus groups has also been shown to enhance social acceptability.~\cite{Kipp2011}.

%Tanja writting
%\newpage
\section{Methodology}
\subsection{Evaluation Design} \label{sec:design}
The evaluation combined summative and formative aspects -- while the implemented functions of the SL avatar were summatively assessed, the findings were also used formatively to support future development. Accordingly, practical testing of the SL avatar application on the Microsoft HoloLens 2 is a core component of the investigation. To address the research questions, a user study was designed with a focus on deaf and hard-of-hearing participants.

To analyze the differences identified in the research questions, the study compared two variants of the SL avatar—one fixed and one adjustable. Technically, both variants are identical. However, in the fixed version, the avatar was displayed in a default position and users were restricted from interacting with it via gesture-based adjustments.

By using a within-subject design, the study captured relative rather than absolute evaluations of the avatar —an important distinction given the current state of avatar development. In the adjustable version (A), participants could modify the avatar’s size, orientation, and spatial positioning using gestures. In the fixed version (F), the avatar remained anchored within the user's field of vision, scaled to a legible size. The version presented first (either F or A) acts as an anchor for the comparative evaluation of the second condition.

To assess comprehensibility rather than overall comprehension, ideally, signs would be presented in isolation and without contextual cues, similar to the ISIT approach described by \cite{Martino2017}. However, many signs—such as pointing or directional signs—derive their meaning from contextual cues within full sentences. Given that the avatar currently only supports playback of pre-stored sentences, the study instead compared comprehensibility across two conditions: one in which participants were given a described situational context (I), and one in which sentences were presented without any contextual information (II). This allowed for an exploration of whether contextual knowledge and prior expectations influence the comprehensibility of avatar-signed content.

This results in the test conditions reported in Table \ref{tab:test_conditions}.

\begin{table}[h]
\centering
\begin{tabular}{p{4.5cm}p{1.5cm}p{1.5cm}}
\toprule
\textbf{Test Condition} & \textbf{Fixed Avatar (F)} & \textbf{Adjustable Avatar (A)} \\
\midrule
Situational comprehensibility (I) & F-I & A-I \\
Pure Comprehensibility (II) & F-II & A-II \\
\bottomrule
\end{tabular}
\caption{Test conditions}
\label{tab:test_conditions}
\end{table}

\begin{figure*}
    \centering
    \includegraphics[width=.7\linewidth]{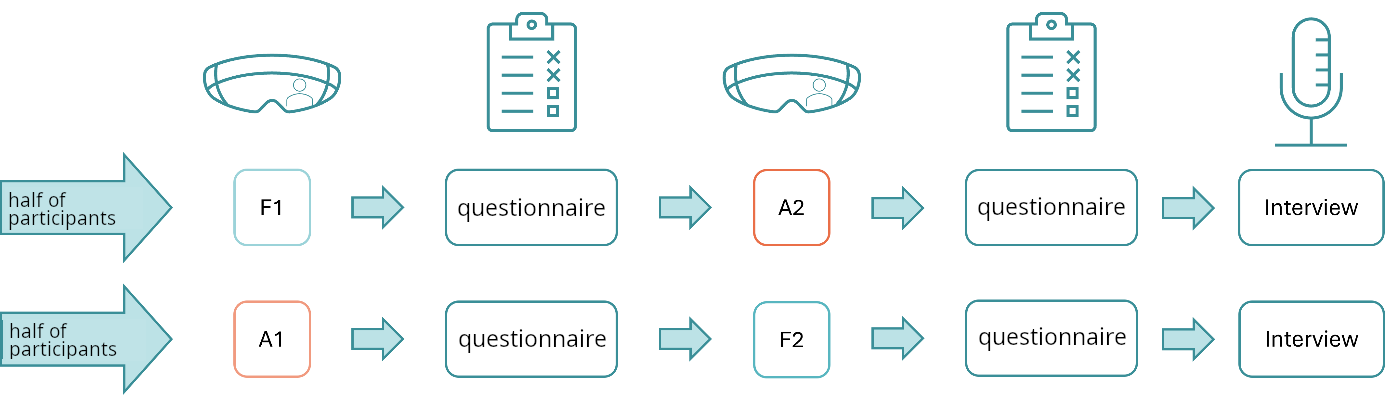}
    \caption{Procedure of the study}
    \label{fig:procedure}
\end{figure*}

In accordance with the previous description of the evaluation design, each participant tested both versions (F and A) of the SL avatar (see Figure \ref{fig:teaser} for a screenshot and Figure~\ref{fig:procedure} for the procedure).

Half of the participants started by testing the fixed avatar (F), while the other half started with the adjustable avatar (A). The order in which participants encountered the situational comprehensibility condition (I) versus the pure comprehensibility condition (II) was also counterbalanced. This variation in sequence was intended to minimize learning and order effects. During the testing phases, objective methods were applied: comprehensibility was assessed using performance-based measures. Additionally, observational data on how participants interacted with the application were collected to provide insights into usability and user experience. Participants' visual perspectives were also recorded via screen capture, enabling retrospective, objective evaluation.

Following each test phase, subjective impressions of user experience and stress levels were collected using standardised questionnaires. These closed-format items facilitate comparison across participants. However, for the formative aspect of the evaluation, qualitative feedback was essential. Therefore, at the end of each session, participants were interviewed using open-ended questions. The combination of qualitative and quantitative data allowed for a comprehensive understanding of user perception, with a particular emphasis on how the system was understood \cite{Wechsung2014}.

Participants could choose whether they preferred the instructions and questionnaires in written form or in SL. If the SL option was selected, the materials were given by the study lead and interpreted by a professional SL interpreter. This ensured that all participants received the same content regardless of format. For participants who chose the written version, individual terms could have been translated on request. SL interpreters had received all materials in advance, along with a written briefing emphasising consistent translation in order to minimise potential bias.

\subsection{Test Phases}

In one of the two test phases, participants used the avatar in its fixed version. In this case, the avatar was permanently displayed above the avatar menu and anchored to the user's field of vision. In the other test phase, participants could use grab-and-drag gestures to adjust the avatar in the way that was most comfortable for them. Instructions were provided in advance of the study, ensuring all participants were familiar with the avatar operation and gestures. If necessary, the study conductor intervened during the adjustable avatar-phase to correct the use of gestures based on the instructions.
To ensure comparable results, all participants interacted with the avatar while standing. The SL interpreter was present as a potential dialogue partner, serving as a point of spatial reference for the adjustments.

Each test phase included three pre-recorded animated signed sentences that participants were to select and start in the avatar menu. For one sentence (test condition I) per test phase, participants were asked to imagine themselves in a given situation, enhancing the evaluation of comprehensibility. They were given a scenario in which, for example, they were at a train station waiting for a train. The more detailed information allowed them to assume an expectation of certain contents of the subsequent sentence. In order to maintain the focus on the evaluation of comprehensibility, the two other sentences (test condition II) originated from the same superordinate scenario, but no further information on the expected content of the sentence was provided. Furthermore, these sentences were not related to each other in terms of content.

%here goes table with sentances 
\begin{table}[h]
\centering
\footnotesize
\begin{tabular}{p{1.6cm}p{2.3cm}p{3.7cm}}
\toprule
\textbf{} & \textbf{Fixed Avatar (F)} & \textbf{Adjustable Avatar (A)} \\
\midrule
\textbf{Situational comprehensibility (I)} & ``Unfortunately, the train is delayed by 15 minutes. It is expected to depart at 1 pm.'' & ``ICE 1557 from Mainz, arrival 8:41 a.m., is canceled today. This is due to a technical fault on the train.'' \\
\midrule
\textbf{Pure Comprehensibility (II)} & ``Arrival RE 77 to Cologne main station via Hanover, departure originally 3:44 a.m. \newline No stop in Cologne today.'' & ``Track 18a, information on ICE 1557 from Mainz, arriving at 13:20, today on track 18b. I repeat: ICE 1557 from Mainz, arrival 13:20, today on track 18b. Please do not board. \newline Arrival RE 77 to Cologne main station via Hanover, departure 3:44 a.m.'' \\
\bottomrule
\end{tabular}
\caption{Assigned sentences according to test conditions}
\label{tab:test_conditions_sentences}
\end{table}

%\subsection{Questionnaires}
%To record the subjective impression of the tested avatar versions immediately after each test phase and to enable comparison of results between all participants, standardized questionnaires are used. The User Experience Questionnaire (UEQ) for measuring the user experience and the Nasa Task Load Index (Nasa TLX) for measuring stress serve as the basis. The UEQ-S short version is used to counteract a lengthy response process and declining response quality.

\subsection{Questionnaires}

To capture the subjective impression of the tested avatar version immediately after each test phase and to facilitate comparison of results between all participants, two standardised questionnaires were employed.

The User Experience Questionnaire (UEQ) was designed to provide a quick and comprehensive impression of the user experience, focusing on the subjective perception of product features and their influence on users \cite{Laugwitz2008}. Its short version, UEQ-S was used for evaluating two versions of the avatar in one session to reduce the length of the response process and prevent fatigue-related declines in response quality.

The use of the avatar's setting options placed additional demands on users. Therefore, recording perceived stress provided insight into overall strain and contributing factors. The NASA TLX is a questionnaire that measured this stress in six dimensions \cite{Hart2006}: mental demand, physical demand, temporal demand, performance, frustration, and effort, each recorded on a continuous scale (0 to 100) and assessed separately in 21 gradations \cite{NASA_TLX}.
For the calculation of total stress, the Raw TLX (RTLX) was used, where the individual dimensions were averaged, allowing for a reduction in response time. Another application of the NASA TLX was to evaluate the individual dimensions instead of the total stress, which was also considered in the evaluation.

The results of the questionnaires aimed to indicate the perceived difference between the two avatar versions. To this end, the results for the fixed avatar were compared with those for the adjustable avatar. Furthermore, a more differentiated approach was chosen: subdividing results according to avatar versions and first and second test phase enabled an evaluation of the unbiased perception in the first test phase. In addition, this differentiation made it possible to evaluate the change in results between the two test phases according to the respective anchor (F/A).  A trend could be identified from the respective mean values. Confidence intervals were also evaluated with a confidence level of 95\% and a one-sided t-test was used to test the tendency of the mean difference for statistical significance.

\subsection{Comprehensibility}
% Intelligibility can be measured subjectively and objectively. There is as yet no method for measuring the intelligibility of SL (Crowe et al., 2019, p. 994). The study shows that the subjective assessment of a signing person correlates with the intelligibility of their signing and an objective observation (Crowe et al., 2019). However, this relates to the signer's own SL production. Another study on comprehension, on the other hand, suggests that the subjective assessment is lower than the actual comprehension (Smith \& Nolan, 2016, p. 572). In order to obtain a differentiated evaluation of comprehensibility in the various sentence components, comprehensibility is measured objectively on the basis of repeated sentences. 

Although users of DGS were involved in our research team, there was not enough capacity to perform a direct analysis of the repeated signed sentence within the scope of this work. 
Therefore, the evaluation was done based on the transcript of the professional SL interpreters who translated the sentences repeated by the participants in DGS into spoken language. 
For the animated sentences of the SL avatar, there was both a transcript of the spoken language and a transcript of the SL in gloss form. There was no direct word-to-sign translation between spoken language and SL.
% (World Federation of the Deaf \& WASLI, 2018). 
As a result, the sentences were segmented into the smallest possible common meaning-bearing unit. This was, for example, a time, a place or a direction. The equivalent glosses and words in spoken language were assigned to this unit. The assessment itself was carried out using a performance measure in both binary and differentiated form. The comprehensibility of a unit was assigned a value. In the binary evaluation, a correctly reproduced meaning was given the value 1, otherwise the unit was given the value 0. The differentiated form was an approach to give SL a higher significance. As individual units may have consisted of multiple glosses, the following evaluation was used in this method:

\begin{table}[h]
\centering
\small
\begin{tabular}{cp{6.5cm}}
\toprule
\textbf{Score} & \textbf{Criterion} \\
\midrule
0 & Gloss not repeated or gloss recognised but no meaning could be assigned to it \\

0.5 & Gloss repeated and several possible meanings were assigned, the correct meaning is below \\

1 & Gloss correctly understood \\
\bottomrule
\end{tabular}
\caption{Evaluation criteria for comprehensibility}
\label{tab:comprehensibility}
\end{table}

If a unit consisted of several glosses describing the meaning of the unit, the score was calculated for each gloss and then averaged. Due to limited resources in DGS expertise, we could not fully validate the quantification, which is why a statistical evaluation of the quantitative results was not carried out. The values represented a guideline for the test conditions under which comprehensibility was better or worse. The binary evaluation served as a control mechanism.
In the further evaluation, all scores were averaged and thus resulted in an average assessment of comprehensibility. Averaging without weighting the individual units was intended to enable a neutral evaluation that avoided bias given our limited resources regarding DGS expertise. Neither communicability nor comprehension was evaluated. Therefore, a statement as to whether the units that are relevant for capturing the content of the sentence were understood is of no further significance for this work.

\subsection{Interview}
To gather subjective, qualitative feedback, an interview was conducted. Its aim was to understand the users' perceptions and explore the causes. In addition, this method could be used to identify approaches for further development and research. 
The interview was structured by three guiding questions, which could then be individually adapted to the answers in greater depth. This method focused on the \textit{user experience} of the settings options and the \textit{acceptability} of an avatar with settings options. At the same time, anomalies from the observation of the interaction or from the answers to the questionnaires were addressed. Finally, the participants had the opportunity to give free feedback.

The qualitative feedback consisted of observations from the session itself and the analysis of the screen recordings as well as participants' interview responses. During the evaluation process, the feedback was first clustered thematically. The aim was to examine the various areas of interaction in the avatar application and to understand the quality of use. In a second step, the importance for the application was evaluated based on findings from previous research and the frequency of occurrence. The results of the questionnaires were then combined with the qualitative feedback.

%Participants here

\subsection{Participants}\label{participants}

Nine individuals participated in the study, all of whom were users of SL and affiliated with the Centre for Deaf Culture and Visual Communication Berlin/Brandenburg (ZFK e.V.)\footnote{\url{https://zfk-bb.de/}}, where the study was conducted. Four participants were aged 20-30, four were 40-50, and age was unreported for one participant.
Seven participants (88.9\%) were deaf; two of them used technical aids, and four regularly relied on SL interpreters. One participant was hearing and worked as a SL interpreter.
Eight participants completed the study in full. One participant was unable to start the animation in the first test phase, leading to the exclusion of their quantitative data, though their qualitative feedback was retained. 
All participants gave a written informed consent prior to the study according to the GDPR requirements, following an approval by the DFKI ethics committee.

\section{Results} %Lefteris 

\subsection{Comprehensibility}\label{comprehensibility-2}

The evaluation of comprehensibility in binary and differentiated form
came to a qualitatively similar result, only the individual values
differ slightly.
It did not suggest that the setting options make a difference to comprehensibility. 
On average, less than 50\% of the individual sentence parts were understood. 
The situational comprehensibility of the fixed avatar was slightly higher than the average comprehensibility of all other sentences.

The transcript of the SL interpreters generally indicates
that it was mainly fragments in the form of individual signs or
phrases that the users managed to repeate, rarely complete sentences. 
In some cases, nothing was understood at all and the subjective perception that the SL avatar was difficult to understand was reported several times.

When analyzing the individual sentence components, it was noticeable that certain parts in particular tended to be more comprehensible on average. 
For example, times and time information were understood quite well and locations were also mostly understandable -- with the exception of "Mainz". 
In the case of locations, however, the lack of a mouthing was particularly evident with "Hannover", since the manual sign for "Hanover" is identical to that for "blue" without a mouthing. In this case, the mouthings would have been essential and correct comprehension could only be possible if sentence context permits. 
This was also observed with other manual signs such as the gloss "PLAN", which was often understood as "technical". 
There were also still gaps in the comprehensibility of times and time units: 
In some cases, users only understood the hour component of the signed times, or parts of the signed numbers, such as "3" instead of "13".
Such examples are obvious when the hand shapes show similarities.

The comprehensibility progression in relation to the chronology of the
sentence shows that the units at the beginning were particularly
difficult to understand. 
This effect is particularly evident with the
adjustable avatar.
% ( Figure7 )
In short sentences, it also happened
that a sentence was completely missed.

% \includesvg[width=6.27668in,height=5.65094in]{./media/image12.svg}

% \protect\phantomsection\label{_Ref181628313}{}Figure7 Evaluation of
% comprehensibility according to sentence units

When differentiating the users based on whether they actually managed to apply the
adjustments, comprehensibility tends to be slightly better on average for those who had actually adjusted the avatar. 
Nevertheless, among them, no increase in comprehensibility from the fixed to the adjustable avatar version could be observed. 
The comprehensibility of this group remains largely constant.
Among those who did not change the avatar, the average comprehensibility of the sentences varied more strongly overall. On average, comprehensibility is slightly lower for the adjustable version than for the fixed avatar version. However, this is more likely due to aspects of individual comprehensibility or the difficulty level of individual sentences, as otherwise the average comprehensibility for both avatar versions should not have changed for this group.

\begin{figure}
    \centering
    \includegraphics[width=1.05\linewidth]{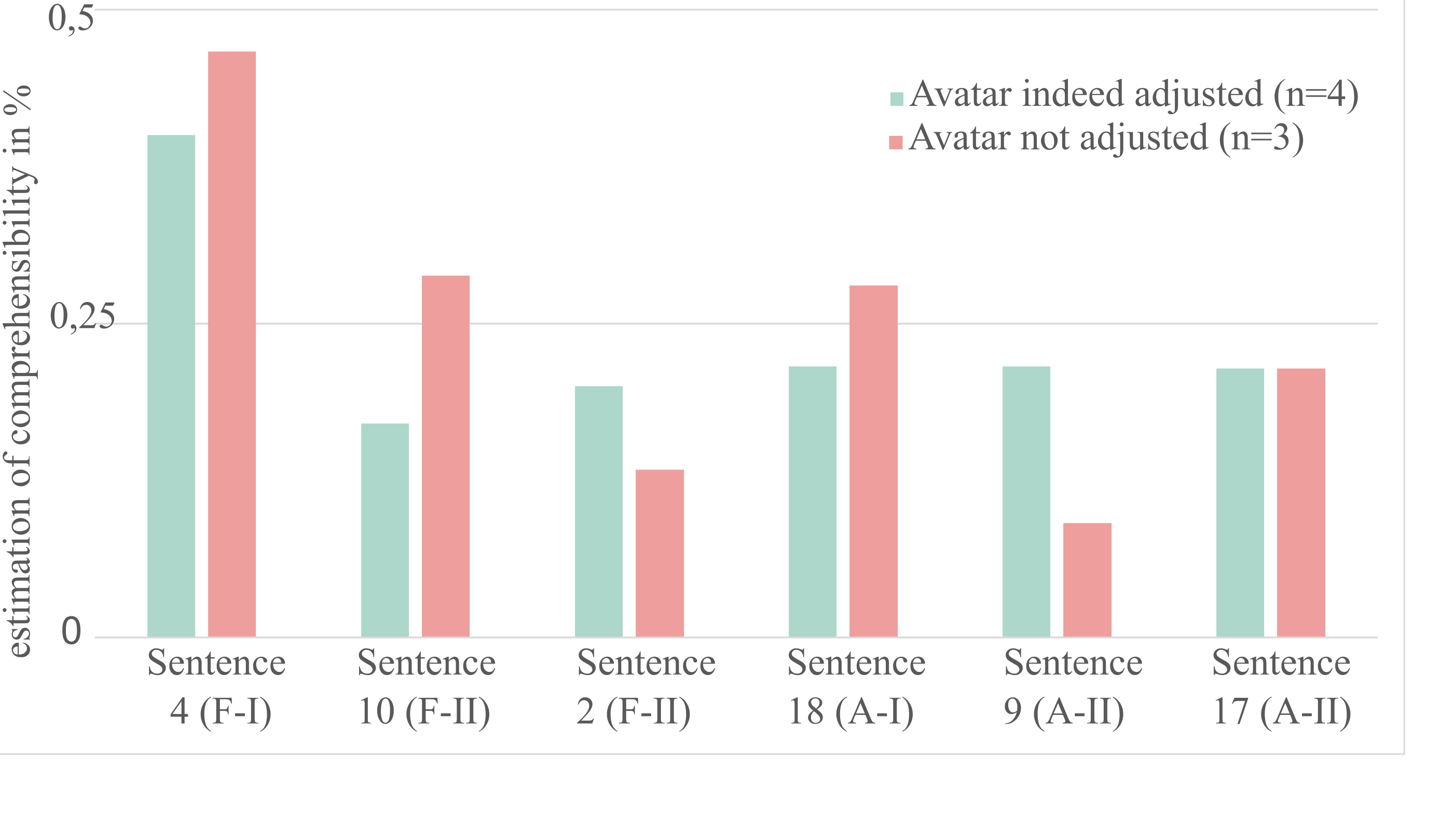}
    \caption{Average comprehensibility after actual adjustment of the avatar. Values of the comprehensibility of the individual sentences after using the setting options}
    \label{fig:graph_comprehensibility}
\end{figure}
% \includesvg[width=5.89623in,height=3.43752in]{./media/image14.svg}

% Average comprehensibility after actual adjustment of the avatar Values of the
% comprehensibility of the individual sentences after using the setting
% options

No clear correlation was found between the existing sign
language levels and comprehensibility. 
However, this observation cannot be generalised, as all participants had a high level of proficiency in DGS. 
An evaluation of the average comprehensibility per participant over the course of the session showed no learning effects that could have influenced the results of comprehensibility.

\subsection{Evaluation of
questionnaires}\label{evaluation-of-questionnaires}

% In the first step, the results are examined for inconsistencies in the
% data. The Short UEQ Data Analysis Tool points out that response patterns
% and non-differentiated responses can occur, especially with the brevity
% of the UEQ-S. Data sets with an inconsistency in both the hedonic
% quality scale and the pragmatic quality scale indicate a suspicion that
% possibly no serious answers were given. In the results of the UEQ-S, an
% inconsistency in the pragmatic and hedonic scale is found for the fixed
% avatar for one data set. The results for the adjustable avatar also show
% a critical data set with inconsistencies in both scales. The participant
% with the critical data set for the adjustable avatar also did not answer
% the efficiency scale for the fixed avatar. However, the Short UEQ Data
% Analysis Tool only recommends removing data sets if all items are
% consistently answered with the middle answer option. This is not the
% case for the available data.

\subsubsection{UEQ-S}\label{user-experience-questionnaire---short}

\begin{figure}
\centering
\includegraphics[width=0.48\textwidth]{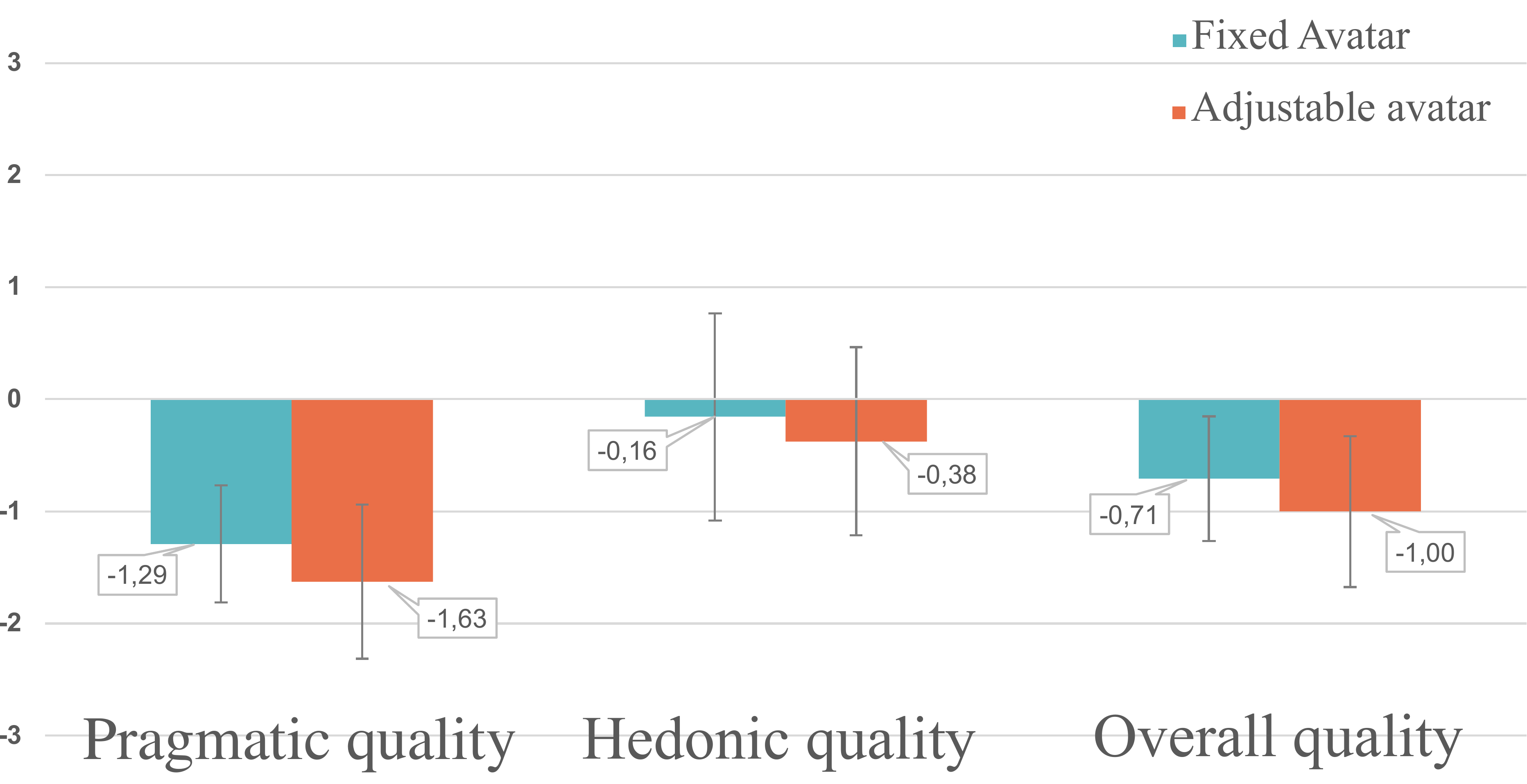}
\caption{UEQ-S, fixed vs. adjustable avatar mean values}
\label{fig:ueq-s-both}
\end{figure}

Both avatar versions received an overall negative user
experience rating (Figure \ref{fig:ueq-s-both}). 
% The fixed avatar (\emph{M\textsubscript{F }}=
% -0.71, KI\textsubscript{F} = {[}-1.27,-0.154{]})\footnote{For better
%   readability, the decimal values are listed in the text with dot
%   notation.} tends to be rated slightly better than the adjustable
% avatar (\emph{M\textsubscript{A }}= -1.00, KI\textsubscript{A} =
% {[}-1.67,-0.33{]}), but no significantly better rating can be determined
% (p = 0.2095).
The fixed avatar was rated slightly better than the adjustable
avatar, although the difference is not statistically significant. 
The hedonic quality of the fixed avatar was rated higher than its pragmatic quality.

% The better results for the fixed avatar are also reflected in the
% differentiated consideration of the pragmatic and hedonic quality. The
% hedonic quality (\emph{M\textsubscript{F }}= -0.16, KI\textsubscript{F}
% = {[}-1.08,0.77{]}; \emph{M\textsubscript{A }}= -0.38,
% KI\textsubscript{A} = {[}-1.21,0.46{]}) is rated better for both
% versions than the pragmatic quality (\emph{M\textsubscript{F }}= -1.29,
% KI\textsubscript{F} = {[}-1.81,-0.77{]}; \emph{M\textsubscript{A }}=
% -1.63, KI\textsubscript{A} = {[}-2.13,-0.94{]}).

\begin{figure}
\centering
\includegraphics[width=0.48\textwidth]{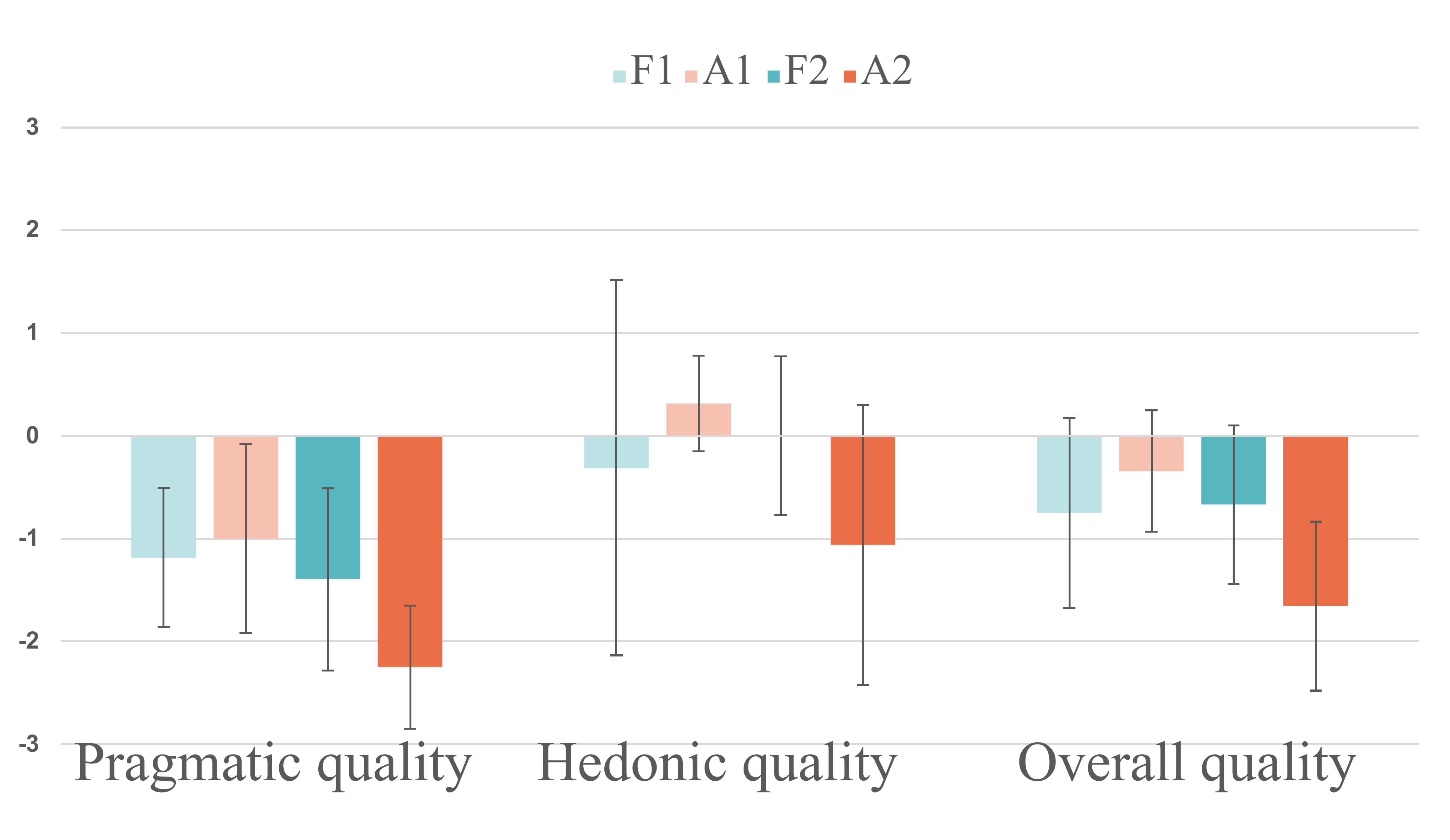}
\caption{\protect\phantomsection\label{_Ref180844380}{}UEQ-S, both phases, mean value comparison}
\label{fig:ueq-s-diff}
\end{figure}

When breaking down results by which version of the avatar was seen first (F1, A1) or second (F2, A2),
the adjustable avatar achieved slightly more positive results than the fixed avatar (Figure \ref{fig:ueq-s-diff}). 
The hedonic quality of the adjustable avatar (A1) was rated significantly better than its pragmatic quality.

The consequent ratings show that the second test 
phase received lower ratings, regardless of the order in which the
avatars were tested. The participants who tested the adjustable avatar
first and then the fixed avatar rated the UX of the fixed avatar
slightly, but not significantly worse. In contrast, the
participants who first tested the fixed and then the adjustable avatar,
rated the UX of the adjustable avatar significantly worse.

% \begin{longtable}[]{@{}
%   >{\centering\arraybackslash}p{(\linewidth - 6\tabcolsep) * \real{0.0688}}
%   >{\centering\arraybackslash}p{(\linewidth - 6\tabcolsep) * \real{0.3558}}
%   >{\centering\arraybackslash}p{(\linewidth - 6\tabcolsep) * \real{0.3261}}
%   >{\centering\arraybackslash}p{(\linewidth - 6\tabcolsep) * \real{0.2493}}@{}}
% \caption{\protect\phantomsection\label{_Toc183358806}{}Table4 Data table
% of the mean values}\tabularnewline
% \toprule\noalign{}
% \begin{minipage}[b]{\linewidth}\raggedright
% \end{minipage} & \begin{minipage}[b]{\linewidth}\raggedright
% \textbf{Pragmatic quality}
% \end{minipage} & \begin{minipage}[b]{\linewidth}\raggedright
% \textbf{Hedonic quality}
% \end{minipage} & \begin{minipage}[b]{\linewidth}\raggedright
% \textbf{Overall quality}
% \end{minipage} \\
% \midrule\noalign{}
% \endfirsthead
% \toprule\noalign{}
% \begin{minipage}[b]{\linewidth}\raggedright
% \end{minipage} & \begin{minipage}[b]{\linewidth}\raggedright
% \textbf{Pragmatic quality}
% \end{minipage} & \begin{minipage}[b]{\linewidth}\raggedright
% \textbf{Hedonic quality}
% \end{minipage} & \begin{minipage}[b]{\linewidth}\raggedright
% \textbf{Overall quality}
% \end{minipage} \\
% \midrule\noalign{}
% \endhead
% \bottomrule\noalign{}
% \endlastfoot
% F1 & -1,19 & -0,31 & -0,75 \\
% A1 & -1,00 & 0,31 & -0,34 \\
% F2 & -1,40 & 0,00 & -0,67 \\
% A2 & -2,25 & -1,06 & -1,66 \\
% \end{longtable}

The results also show that if the users had actually adjusted the avatar, the user experience of the adjusted avatar was rated worse. 
In this subgroup, the perception of pragmatic quality was further apart between the two avatar versions than in hedonic quality.

\subsubsection{RTLX}\label{raw-task-load-index}

\begin{figure}
\centering
\includegraphics[width=.5\textwidth]{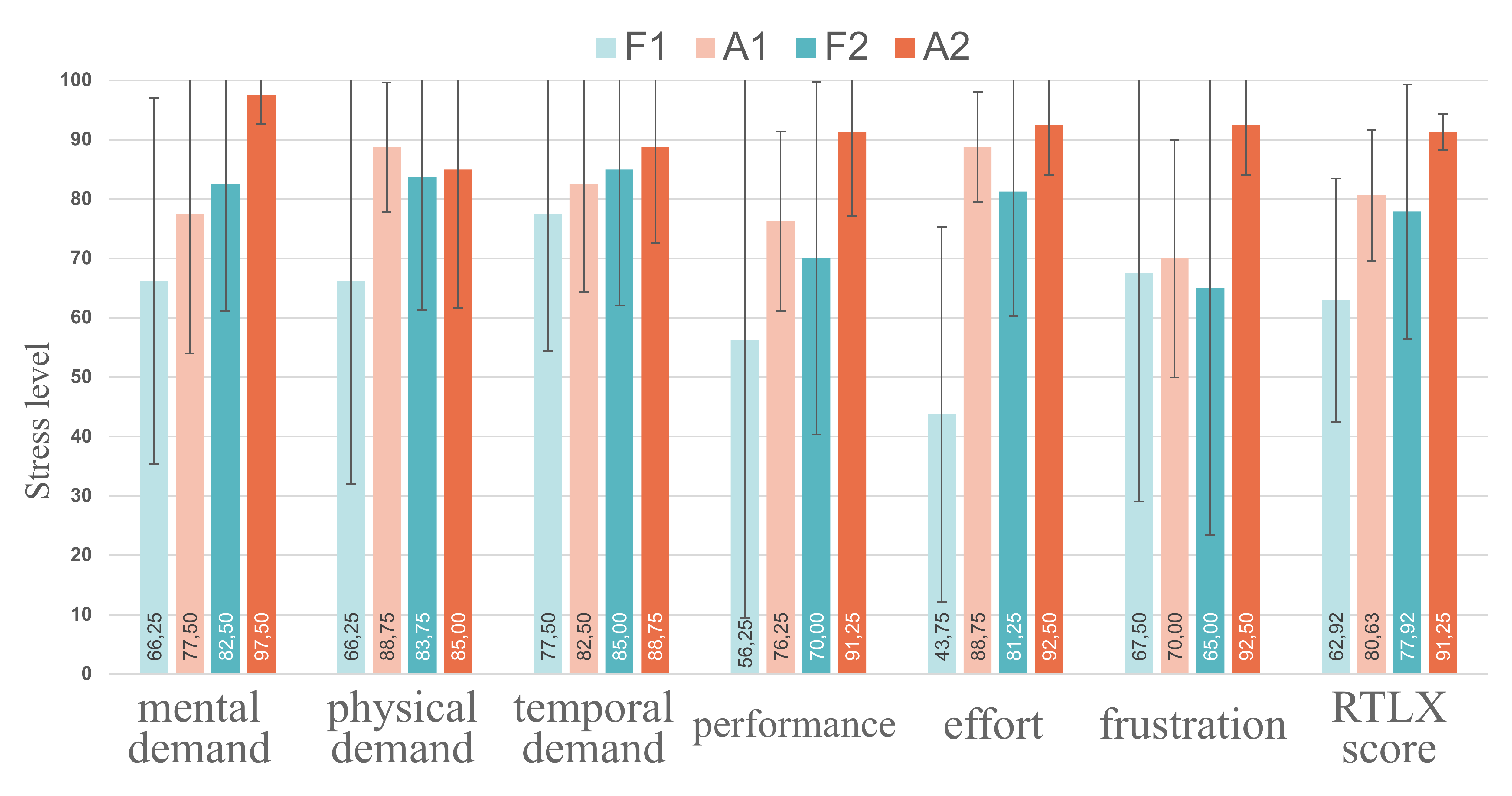}
\caption{\protect\phantomsection\label{_Ref180844542}{}RTLX,
differentiated mean value comparison}
\label{fig:rtlx-diff}
\end{figure}

The generally lower UX for the adjustable avatar is also reflected in
the results for stress (Figure \ref{fig:rtlx-diff}). The average stress of the
adjustable avatar was rated significantly higher than that of the
fixed avatar. 
% (\emph{M\textsubscript{A }}= 85.94; \emph{M\textsubscript{F}}= 70.42; t = -1.97, df = 7, p = 0.0448)
The individual dimensions show that the biggest
differences between the avatar versions were found in effort, performance and
frustration. However, only effort was significantly higher for the
adjustable avatar.

% (\emph{M\textsubscript{AN-F }}= 62.50;
% \emph{M\textsubscript{AN-A }}= 90.63; t = -2.32, df = 7, p = 0.027), 
% the
% differences in the dimensions performance and frustration describe a
% tendency. 
% In the present results, effort is the most stressful dimension
% for the adjustable avatar, whereas it makes the smallest contribution
% to stress for the fixed avatar.
% 
It is also noticeable that the confidence intervals for the fixed avatar
were larger than for the adjustable version,
% For the fixed avatar, they
% sometimes include half of the scale. 
% The reason for this 
% lies in the variance of the answers: 
since the distribution of answers was more
scattered for the fixed version than for the adjustable version. 
The frequency distribution of the responses in the individual
dimensions shows that participants assessed the adjustable avatar
exclusively with stress values greater than 50, 
% (SB\textsubscript{A} =
% 50, min\textsubscript{A} = 50, max\textsubscript{A} = 100), 
while ratings cover the full range of the scale for the assessment of the fixed avatar. 
% (SB\textsubscript{F} = 100, min\textsubscript{F} = 0,
% max\textsubscript{F} = 100). 
This indicates that the assessment for the
adjustable avatar was more uniform than for the fixed avatar, 
whereas the distribution of the values across the entire scale and
the use of the various gradations indicate that the assessment was
fundamentally differentiated.

% \begin{figure}
% \centering
% \includesvg[width=5.47917in,height=3.21042in]{./media/image20.svg}
% \caption{\protect\phantomsection\label{_Ref180844420}{}Figure11 Results
% of the RTLX, comparison F and A}
% \end{figure}

The results of the participants who started with the fixed avatar
% (\emph{M\textsubscript{F1}} = 62.92, \textsubscript{KI(F)1} =
% {[}42.37,83.47{]}) 
show the lowest mean value, indicating the lowest perceived stress (Figure \ref{fig:rtlx-diff}). Except for frustration, all other dimensions are the lowest
compared to other test conditions. The effort dimension
% (\emph{M\textsubscript{AN-F1 }}= 43.75) 
shows the most pronounced
difference when comparing the first impression and the change to the
adjustable avatar.
When looking at the versions tested first, the adjustable avatar
is associated with slightly higher perceived stress, but the difference is not
significant
% (\emph{M\textsubscript{F1 }}= 62.92;
% \emph{M\textsubscript{A1 }}= 80.63; p = 0.101).

The analysis of the change in the perception of stress (RTLX score) between the fixed and the adjustable avatar shows a
consistent picture: 
the lower perception of stress of the fixed avatar is independent of the
order presented. 
Nevertheless, the adjustable avatar
showed a significantly higher stress level only when
the fixed avatar was tested first. 
% (\emph{M\textsubscript{F1}} = 62.92;
% \emph{\textsubscript{M(A)2}} = 91.25; t = -2.46, df = 3, p = 0.0453). 
If
the adjustable avatar was tested first, the perceived stress decreased when switching to the fixed avatar, but not significantly.
% (\emph{M\textsubscript{A1 }}=80.625; \emph{M\textsubscript{F2 }}= 77.92
% ; p = 0.148)
The adjustable avatar thus shows the highest average
stress levels in the second test phase. In addition, frustration levels appear particularly high for the adjustable avatar in the second
test phase.

The analysis of the individual dimensions shows that the mental demand
and the time demand could have increased as the test phases progressed.
On the other hand, the remaining dimensions reflect the possible
increased stress caused by the avatar's adjustment
options.

The results of the participants who had actually adjusted the avatar
indicate the greatest difference between the fixed and adjustable avatar
in perceived effort. Both groups rated the adjustable avatar as
more strenuous to use. The frustration dimension shows a fairly large
difference between the fixed and adjustable avatar for those who had
actually adjusted it.

\subsection{Qualitative results}\label{qualitative-results-of-the-interaction-with-the-avatar}

The observations and interviews can be summarised in three areas with
regard to usability: the interaction feedback of the application, the
controllability of the avatar and the adjustment options. 

\subsubsection{Interaction feedback from the
app}\label{interaction-feedback-from-the-application}

From the observation and the reactions of the participants, it can be
deduced that the interaction feedback provided by the application was
often not perceived or even entirely absent.
When selecting and playing the sentences, many participants struggled to recognise whether they had successfully selected a sentence
and repeatedly tapped the button even though the sentence was already
loading. The ``\emph{Loading}'' message was often overlooked, 
and some assumed the system had not responded,
even though the selection was successful.

A similar phenomenon was observed when starting the animation: Users
continued to focus on the ``\emph{Play animation}'' button or the menu
instead of looking at the avatar figure. Expecting the avatar to start
signing immediately after selecting the button, users then checked the
state of the avatar themselves. Not every attempt to tap the "\emph{Play
animation}" button worked, so it often took several attempts, which was described as ``stressful''. 
If the avatar figure and the menu were not displayed together in the field
of view, the required head movements intensified this effect. Focusing on the
menu for too long while waiting for feedback led to the beginnings of
sentences or even entire sentences being missed.

Even when the participants adjusted the avatar, they lacked adequate
feedback. Despite having received prior instructions on the gestures,
the participants were often unsure at which adjustment points the avatar
could be manipulated and whether the attempted gestures were correct. The uncertainty was intensified when the application of a
theoretically correct gesture did not work.

\subsubsection{Controllability of the
avatar}\label{controllability-of-the-avatar}

The use of gestures led to difficulties not only with regard to
feedback, but also in other situations. Some participants used gestures
other than those defined in the instructions. Wrist rotations to adjust orientation and zoom gestures familiar from touchscreens to adjust
the size were observed. When asked, some of these were described as more
intuitive. In some cases, however, they were also tried as an
alternative when the instructed gestures failed to work. The
non-functioning of gestures seems to be a recurring problem: Overall, correct use of the gestures proved difficult, with some
participants describing it as "overwhelming". The screen recordings show
that users sometimes applied the gestures before their hand reached the
adjustment points. Other times, the application did not work for unknown
reasons or the adjustment could not be completed. This manifested itself
in the avatar "getting lost" in the movement, even though the gripping
gesture had not been released. It is possible that the gesture or
hand recognition failed at these moments, but this cannot be evaluated
from the available data. Nevertheless, it was noticeable that gestures
had to be used several times -- this was also reported back
accordingly. Some participants also gave up trying to adjust the avatar
using gestures after a certain time, even though the avatar had not yet been optimally adjusted to their preferences.

\begin{comment}
Some participants suddenly saw an unexpected red beam from their hands
in the field of vision of the AR glasses. With these, the necessary
gestures such as tapping or grasping were no longer possible. At the
same time, the participants did not know how to turn this beam off again
at that moment. This often led to the need for assistance.
\end{comment}

\subsubsection{Adjustment options}\label{adjustment-options}

When positioning the avatar, the participants grabbed it and moved it to
position it within arm\textquotesingle s reach. While the
avatar is displayed at a reduced size as long as it is gripped, it
"stands" on the floor once released. This natural behavior
sometimes led to a significant enlargement of the avatar, which was then
perceived as a combination of "too large" and "too close". In order to
place the avatar in a more distant position, users would have had to move
themselves and then step back after positioning the avatar. However, no user
recognised and applied this.

\subsubsection{Quality of the
presentation}\label{quality-of-the-presentation}

Although the participants were informed about limitations such as the
lack of mouthings, as the quality of the SL avatar was
not the focus of the evaluation, participants still explicitly commented on
this issue. The consistent facial expressions were described as
unpleasant and severely impaired the perception of the avatar. This
confirms and highlights the importance of the quality of the avatar and
the animation. Its rigid appearance and lack of individual style were
criticised as well. During the test phases, an unnatural conversational posture
was observed in some participants. In some cases, they leaned forward
slightly when trying to understand the sentences or squinted their eyes, which did not occur during communication with the
interpreter. Three participants also identified the
avatar\textquotesingle s legs as unnecessary and would have preferred to
see only the signing space.

\subsection{Expressed need for the adjustment features}\label{need-for-setting-options}

When asked about whether the adjustment features are needed -- assuming they were easy to use --, 56\% preferred the
adjustment features, 22\% were undecided and 22\% preferred
a non-adjustable avatar. 
However, some participants noted
that it was difficult to answer the question based on the present avatar
due to its usability problems, as they had to refer only to their imagination.

\section{Discussion and Conclusion}

This study examined the comprehensibility, user experience, and acceptability of adding adjustment options on an SL avatar operating on Microsoft HoloLens~2. 
Although the majority of the users expressed a preference for the adjustment features, amidst a lot of technical problems and missing features, no improvements were observed in comprehensibility and user experience. 

The experiments indicated low comprehensibility irrelevant of the adjustment features, mostly due to lack of mouthings and facial expressions, and difficulty of distinction between similar hand shapes. 
The adjustable version did not achieve any significant difference in comprehensibility as compared to the fixed version. It is unclear if this is because the adjustment features are not a requirement for basic understanding, or that the technical limitations were so strong that they overshadowed every other improvement. Suggestions for improvements included better interaction feedback, animated facial features, and a replay function.

For user experience, the adjustable avatar did not perform better overall than the fixed one. Challenges with gesture interaction, unclear feedback, and high effort contributed to lower UX ratings. Pragmatic quality was generally rated lower than hedonic quality, indicating that users found the system more emotionally or aesthetically pleasing than functionally useful. Stress levels were higher for the adjustable avatar, reflecting increased effort, performance and frustration levels. However, the adjustable avatar shows significantly higher stress levels only when the fixed avatar was tested first. UX tends to decline as the session progressed, likely due to repeated usability issues. The concern about whether the adjustment gestures are intuitive, and how easily users could get familiar with them, was raised. Recommendations included visual aids, tutorials, and cropping the avatar to the signing space.

In terms of acceptability, the concept of the adjustment options was preferred by the majority, but its practical value is strongly linked to usability. Poor usability leads to frustration, thereby reducing acceptance. While some concerns were raised regarding accessibility for less tech-savvy individuals, the adjustment options were seen as valuable if they were easy to use. More critically, the general acceptability of SL avatars depends heavily on animation quality and social perception—particularly concerns about avatars replacing human interpreters.

%\section{Summary of Limitations}\label{summary-limitations}
Due to the high cost of interpretation and communication barriers leading to a slow experiment speed (one hour per participant), the study had to be confined to a small sample size.
This is its main limitation, as it reduces the statistical significance and affects the reliability of the quantitative results.
Additionally, while necessary, interpreter involvement introduces variability that cannot be fully controlled. Lastly, the sentence material could not be customised for evaluation purposes due to system constraints.

%\section{Conclusion and Outlook}\label{summary-conclusion}
The study confirms that SL avatar development is promising, especially with personalisation features, but highlights that usability and animation quality are key. Adjustment options alone do not guarantee improved user experience or acceptability. Importantly, comprehensibility should not rely on adjustability -- avatars must be understandable by default. Future development should prioritise facial animation, better interaction design, and participatory development with deaf communities. %Broader evaluations in real-world contexts are needed to confirm these findings and guide application-specific development.

\begin{acks}
The research reported in this paper was primarily conducted as a BSc thesis at the Technical University of Berlin in co-operation with the German Research Center for Artificial Intelligence (DFKI) supported by BMBF (German Federal Ministry of Education and Research) via the project SocialWear (grant no.~01IW20002) and by the European Union via the project SignReality, as part of financial support to third parties by the UTTER project (Horizon Europe, GA: 101070631).
\end{acks}

\bibliographystyle{ACM-Reference-Format}
\bibliography{references,AR}
\end{document}